\documentclass[]{matterlab}

\usepackage[utf8]{inputenc} %
\usepackage[T1]{fontenc}    %
\usepackage{url}            %
\usepackage{booktabs}       %
\usepackage{amsfonts}       %
\usepackage{nicefrac}       %
\usepackage{microtype}      %
\usepackage{xcolor}         %

\usepackage{amsmath,amsfonts,bm}

\def\eqref#1{equation~\ref{#1}}

\def\1{\bm{1}}

\def\vzero{{\bm{0}}}

\def\va{{\bm{a}}}

\def\vh{{\bm{h}}}

\def\vs{{\bm{s}}}

\def\vx{{\bm{x}}}
\def\vy{{\bm{y}}}
\def\vz{{\bm{z}}}
\def\veps{{\bm{\varepsilon}}}

\def\mI{{\bm{I}}}

\DeclareMathAlphabet{\mathsfit}{\encodingdefault}{\sfdefault}{m}{sl}
\SetMathAlphabet{\mathsfit}{bold}{\encodingdefault}{\sfdefault}{bx}{n}

\def\gM{{\mathcal{M}}}
\def\gN{{\mathcal{N}}}

\def\gU{{\mathcal{U}}}

\def\sN{{\mathbb{N}}}

\newcommand{\E}{\mathbb{E}}

\newcommand{\R}{\mathbb{R}}

\usepackage{bm}
\usepackage{parskip}

\usepackage{tikz}
\usepackage{amsmath}
\usepackage{makecell}
\usepackage[version=3]{mhchem}

\usepackage{algorithm}
\usepackage[noend]{algpseudocode}
\usepackage{algorithmicx}

\usepackage{multirow}
\usepackage{placeins}

\renewcommand{\cite}{\citep}

\newcommand{\spm}[1]{{\scriptstyle \pm {#1}}}

\newcommand{\ndiff}{N_{\mathrm{diff}}}

\newcommand{\sdata}{\sigma_{\mathrm{data}}}

\newcommand{\nexta}{p_{\mathrm{type}}}
\newcommand{\nextx}{p_{\mathrm{pos}}}

\newcommand{\smax}{\sigma_\mathrm{max}}
\newcommand{\smin}{\sigma_\mathrm{min}}

\newcommand{\name}{\textsc{Quetzal}}

\title{Scalable Autoregressive 3D Molecule Generation}

\author[1,2,3]{Austin H. Cheng}
\author[4]{Chong Sun}
\author[1,2,3,5,6,7,8,9]{Alán Aspuru-Guzik}

\affiliation[1]{Department of Chemistry, University of Toronto, Toronto, ON, Canada}
\affiliation[2]{Department of Computer Science, University of Toronto, Toronto, ON, Canada}
\affiliation[3]{Vector Institute for Artificial Intelligence, Toronto, ON, Canada}
\affiliation[4]{Department of Chemistry and Chemical Biology, Rutgers University, Piscataway, NJ, USA}
\affiliation[5]{Department of Materials Science \& Engineering, University of Toronto, Toronto, ON, Canada}
\affiliation[6]{Department of Chemical Engineering \& Applied Chemistry, University of Toronto, Toronto, ON, Canada}
\affiliation[7]{Senior Fellow, Canadian Institute for Advanced Research (CIFAR), Toronto, ON, Canada}
\affiliation[8]{Acceleration Consortium, Toronto, ON, Canada}
\affiliation[9]{NVIDIA}

\metadata[Code]{\url{https://github.com/aspuru-guzik-group/quetzal}}

\abstract{

Generative models of 3D molecular structure play a rapidly growing role in the design and simulation of molecules.
Diffusion models currently dominate the space of 3D molecule generation, while autoregressive models have trailed behind.
In this work, we present \name{}, a simple but scalable autoregressive model that builds molecules atom-by-atom in 3D.
Treating each molecule as an ordered sequence of atoms, \name{} combines a causal transformer that predicts the next atom’s discrete type with a smaller Diffusion MLP that models the continuous next-position distribution.
Compared to existing autoregressive baselines, \name{} achieves substantial improvements in generation quality and is competitive with the performance of state‑of‑the‑art diffusion models.
In addition, by reducing the number of expensive forward passes through a dense transformer, \name{} enables significantly faster generation speed, as well as exact divergence-based likelihood computation.
Finally, without any architectural changes, \name{} natively handles variable-size tasks like hydrogen decoration and scaffold completion.
We hope that our work motivates a perspective on scalability and generality for generative modelling of 3D molecules.

}

\begin{document}

\maketitle

\section{Introduction}

Generative models of 3D molecular structure are accelerating the design and simulation of molecules, with applications across chemistry, biology, and materials science \cite{abramson2024accurate,watson2023novo, zeni2025generative}.
Diffusion-based approaches are the prevailing standard \cite{hoogeboom2022equivariant, song2024equivariant, zhang2024symdiff, joshi2025all}, but they typically operate on fixed-size input/output and are computationally intensive to sample from.
In contrast, autoregressive models of 3D molecules have lagged behind in generation quality \cite{gebauer2019symmetry, luo2022autoregressive, daigavane2023symphony, flam2023language, gao2024tokenizing}.
However, autoregressive models offer several compelling advantages: they support arbitrary-size generation, enable exact likelihood computation, and offer potentially faster generation.
Moreover, molecules have a natural tokenization in terms of atoms, which aligns with the paradigm of autoregression.

This performance gap is often attributed to the assumption that diffusion models are suited for continuous spatial data, whereas autoregressive models are designed for discrete domains like text.
Indeed, prior autoregressive methods for 3D structure typically \emph{discretize} coordinates into 3D grids or tokenized \texttt{.xyz} files, discarding important information about spatial continuity.
Recent work by \citet{li2024autoregressive} challenges this assumption by introducing a Diffusion Loss, which jointly trains a lightweight, per-token diffusion model with an autoregressive transformer.
This hybrid architecture enables autoregressive generation of continuous-valued tokens while retaining the scalability of transformers.

In this work, we adopt and extend this idea for 3D molecule generation.
We propose \name{}, a simple yet scalable autoregressive model that generates molecules atom-by-atom, predicting each atom's discrete type and continuous 3D position.
\name{} combines a causal transformer with a small Diffusion MLP to model the position of the next atom, conditioned on the current prefix structure.
This simple design enables \name{} to scale, achieving generation quality that surpasses all autoregressive baselines and competes with state-of-the-art diffusion models, while also significantly improving generation speed.
Furthermore, \name{} natively supports flexible generation tasks such as hydrogen decoration and scaffold completion, which are cumbersome to implement with fixed-size diffusion models.

By revisiting autoregression through the lens of modern scaling but with a continuous spatial inductive bias, \name{} repositions autoregressive models as a competitive and versatile approach for 3D molecule generation.
Our contributions are:

\begin{itemize}
    
    \item We introduce \name{}, an autoregressive model for 3D molecule generation that sequentially predicts the next atom's discrete type and continuous 3D coordinates.

    \item We demonstrate that \name{} outperforms previous autoregressive approaches and competes with the performance of state-of-the-art diffusion models on QM9 and GEOM, while significantly reducing generation time.

    \item We show that \name{} enables novel capabilities over diffusion models, including exact divergence-based likelihood computation and arbitrary-size generation for hydrogen decoration and scaffold completion.
\end{itemize}

\begin{figure}
\centering
\centerline{\includegraphics[width=\textwidth]{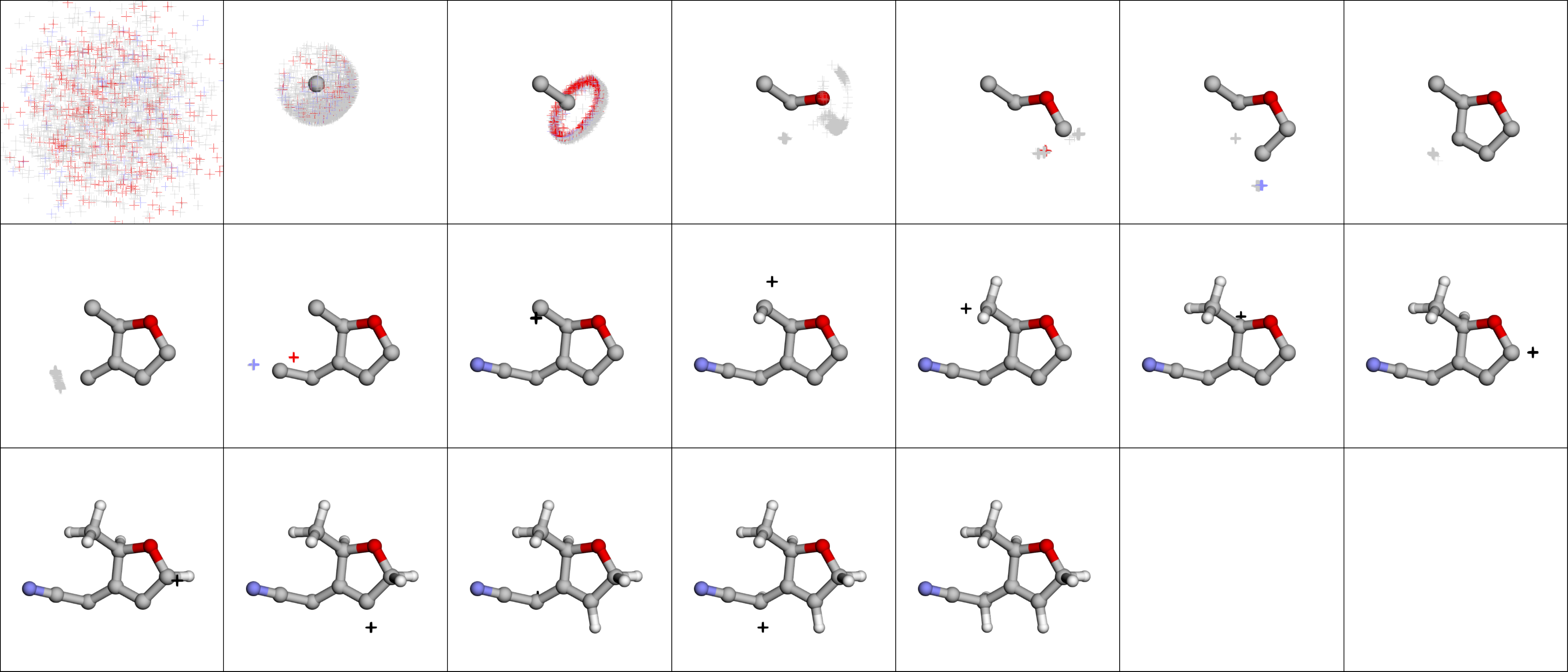}}
\caption{\name{} generates 3D molecules by iteratively predicting the next atom's discrete type and continuous position. Cross marks indicate the distribution of the next atom's type and position.}
\label{seq_grid}
\end{figure}

\section{Related Work}

\textbf{3D molecular generative models.}
Generative models of 3D molecules have been proposed using normalizing flows \cite{garcia2021n}, diffusion models \cite{hoogeboom2022equivariant}, flow matching \cite{song2024equivariant}, Bayesian flow networks \cite{song2024unified}, and latent diffusion \cite{xu2023geometric,joshi2025all}.
Further works have enhanced generation capability by leveraging representation conditioning \cite{li2024geometric} or optimal transport \cite{hong2024fast}.
These approaches are often designed around equivariant architectures and typically model molecules as unordered point clouds.
Other representations such as voxel grids \cite{o20243d} and neural fields \cite{kirchmeyer2025score} have also been explored, which move beyond fixed-size generation.

\textbf{Autoregressive 3D molecular generative models.}
Autoregressive models such as G-SchNet \cite{gebauer2019symmetry} and Symphony \cite{daigavane2023symphony} discretize 3D coordinates and predict relative positions using equivariant architectures.
Other models \cite{luo2022autoregressive, liu2022generating} predict continuous quantities using normalizing flows or mixture models, but remain tied to reference frames.
Another line of work simply casts 3D generative modelling as discrete language modelling of raw \texttt{.xyz} files \cite{flam2023language, gruver2024fine, zholus2024bindgpt, gan2025large} or using custom tokenized representations \cite{wang20253dsmiles, gao2024tokenizing}.

\textbf{Autoregression over continuous-valued tokens.}
The most directly related work to ours is masked autoregression (MAR) \cite{li2024autoregressive}, which introduces a Diffusion Loss for continuous-valued per-token generation in tandem with an autoregressive transformer backbone.
Other approaches such as TimeGrad \cite{rasul2021autoregressive} and Diffusion Forcing \cite{chen2024diffusion} apply similar prefix-conditional, per-token diffusion models for generating continuous-valued sequences.
Instead of per-token diffusion, Jetformer predicts per-token Gaussian mixture parameters \cite{tschannen2024jetformer}, and in this way trains a normalizing flow that understands text and images in data space.
Trans-dimensional jump diffusion enables a  diffusion model to add new dimensions during generation, which resembles autoregression \cite{campbell2024trans}.

\section{Method}

\begin{figure}
\centering
\centerline{\includegraphics[width=\textwidth]{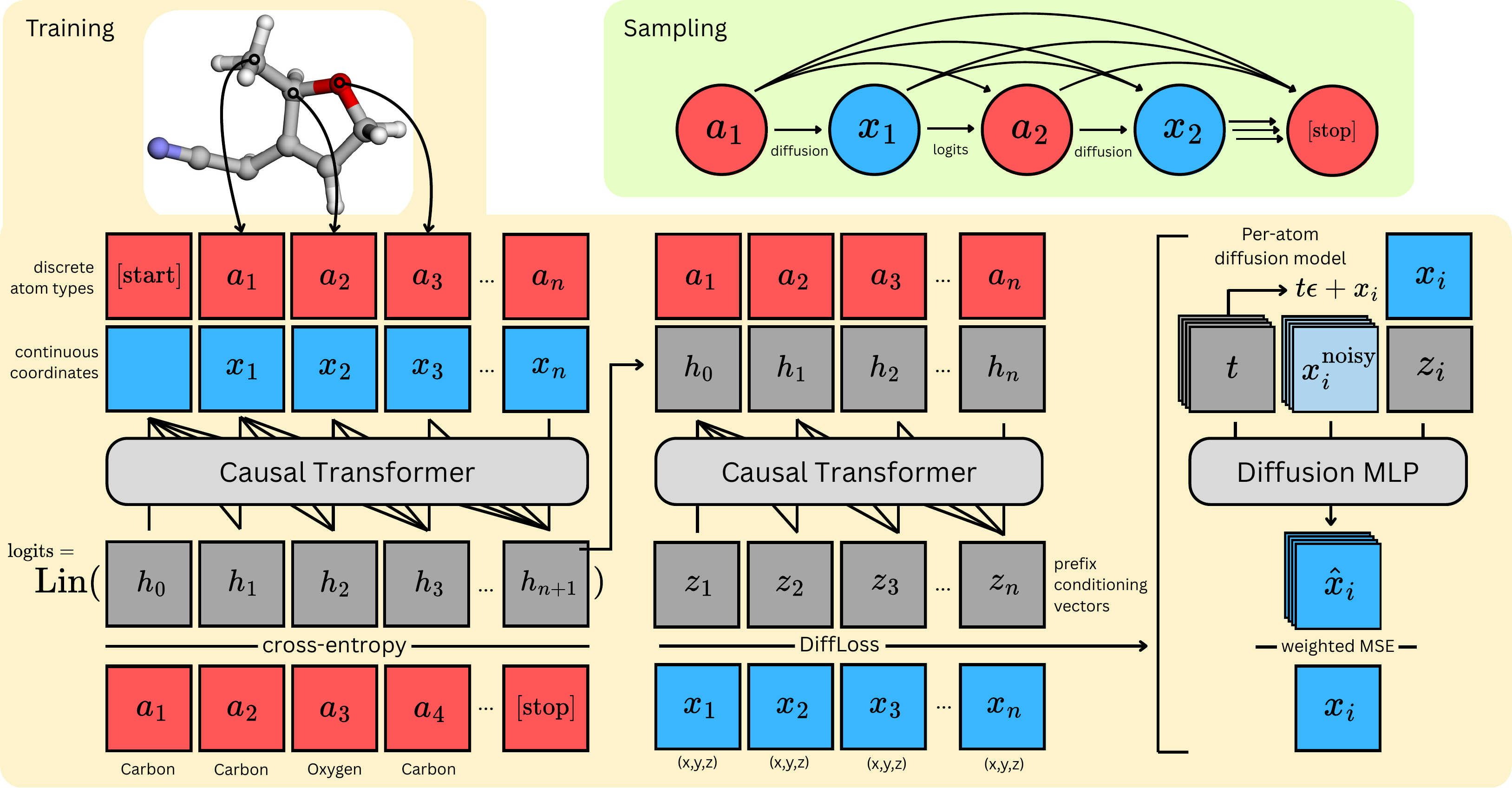}}
\caption{Architecture of \name{} during training and sampling. The first transformer stack causally processes each prefix of an input structure to predict logits of the next atom type. The second transformer stack incorporates information of the next atom type and causally produces conditioning vectors for the next 3D position. The prefix-conditional, per-token Diffusion MLP is trained jointly with the transformer stack using multiple timesteps. Simultaneously batching across length and time provides dense supervision signal. Sampling iteratively predicts atom type and 3D position until \texttt{[stop]} is predicted.}
\label{schematic}
\end{figure}

We consider a 3D molecule $\gM=(\va,\vx)$ as an \emph{ordered} sequence of atom types $\va = (a_1, ..., a_n) \in \sN^{n}$ and coordinates $\vx = (x_i, y_i, z_i)_{i=1}^n = (\vx_1, ..., \vx_n) \in \R^{n \times 3}$, where the molecule contains $n$ atoms.
We tokenize sequences into atoms, so we refer to atoms and tokens interchangeably.
We denote $\vx_{:i} = (\vx_1, ..., \vx_i)$ as the \emph{prefix} of $\vx$ at sequence index $i$, which contains the current token and all previous tokens.
We refer to $(\va_{:i},\vx_{:i})$ as the \emph{prefix structure}.
We index from 1, so $(\va_{:0},\vx_{:0})$ is just the start token.

We define an autoregressive model of 3D molecules which iteratively predicts the next atom's type and position $(a_{i+1},\vx_{i+1})$ given a prefix structure $(\va_{:i},\vx_{:i})$, as shown in \Cref{seq_grid}.
\begin{align}
    p(\gM) &=
     \prod\limits_{i=0} p(a_{i+1},\vx_{i+1}|\va_{:i},\vx_{:i}) = \prod\limits_{i=0} p_{\mathrm{type}}(a_{i+1}|\va_{:i},\vx_{:i})p_{\mathrm{coord}}(\vx_{i+1}|\va_{:i+1},\vx_{:i})
\end{align}
The model alternates between predicting $a_i$ and $\vx_i$ (i.e. sampling from $\nexta$ and $\nextx$) until a \texttt{[stop]} token is predicted, or until a maximum number of atoms is reached.
The \texttt{[stop]} token is considered as an atom type, so $\nexta$ is also responsible for modelling whether to stop generation.
The generative model is fully specified when we specify these two conditional distributions.
The next-type distribution $\nexta(a_{i+1}|\va_{:i},\vx_{:i})$ is a categorical distribution.
This is straightforward to model using a standard GPT.
First, we embed the prefix structure $(\va_{:i},\vx_{:i})$ using embeddings for the atom types, linear layers and Fourier encodings \cite{tancik2020fourier} for the coordinates, and learned positional encodings for sequence ordering.
The combined embeddings are then passed through a causal transformer \cite{vaswani2017attention}:
\begin{equation}
    \vh_{i} = \mathrm{Transformer}\left(\mathrm{Emb}(\va_{:i})+\mathrm{Lin}(\vx_{:i})+ \mathrm{Lin}(\mathrm{Fourier}(\vx_{:i})) +\mathrm{PosEmb}_{:i}\right).
\end{equation}
Because attention is causally masked, the causal transformer produces prefix embeddings $\vh_i$ for all prefixes in the sequence in one forward pass (\Cref{schematic}).
Each prefix embedding $\vh_{i}$ is then passed to a linear layer with no bias which predicts logits of the next atom type,
\begin{equation}
    p(a_{i+1}|\va_{:i},\vx_{:i}) = p(a_{i+1}|\vh_{i}) = \mathrm{softmax}(\mathrm{Lin}(\vh_{i})),
\end{equation}
which are supervised by cross-entropy loss against the ground truth next atom types.

What remains is to model the next 3D position distribution $\nextx(\vx_{i+1}|\va_{:i+1},\vx_{:i})$.
Here, instead of discretizing $\vx$, we use the Diffusion Loss proposed by \citet{li2024autoregressive} to model the continuous distribution $\nextx$.
In other words, we model $\nextx$ as a diffusion model that generates a single 3D position, conditioned on a prefix structure and next atom type.
We first combine the prefix embeddings $\vh_{:i}$ with the next atom type $\va_{i+1}$ and pass through a second transformer to obtain the single-atom conditioning vector $\vz_i$, which conditions the diffusion model:
\begin{align}
    \nextx(\vx_{i+1}|\va_{:i+1},\vx_{:i}) = p(\vx_{i+1}|\vz_{i+1}), \; \textrm{where} \; \vz_{i+1} = \mathrm{Transformer}(\mathrm{Emb}(\va_{:i+1})+\vh_{:i}).
\end{align}
This structure forces the model to commit to a discrete atom type before predicting its continuous position, which we find is beneficial for learning.

We now introduce diffusion models, and temporarily drop the subscripts and the conditioning vector.
Our approach closely follows \citet{karras2022elucidating}.
Diffusion models learn to sample from a continuous distribution defined by a dataset $p_{\mathrm{data}}(\vx)$.
We can corrupt data by taking every data example and adding i.i.d. Gaussian noise with standard deviation that scales linearly with time $t$.
This defines a time-dependent probability density $p_t(\vx)$ described by convolving the data distribution with a Gaussian whose standard deviation is linearly increasing in time, $p_t(\vx) = p_{\mathrm{data}}(\vx) \otimes \gN(\vzero, t^2 \mI)$.
At small times, $p_0$ approximates the data distribution, whereas for large time $T$, $p_T$ is well approximated as a large Gaussian $\gN(\vzero, T^2 \mI)$, which can be sampled without knowing $p_{\mathrm{data}}$.
Then, a remarkable result is that samples $\vx_t \sim p_t(\vx_t)$ can be generated by drawing samples $\vx_T\sim p_T(\vx_T)$ and evolving them backwards from time $T$ to $t$ under the probability flow ODE \cite{song2020score, karras2022elucidating},
\begin{equation}
\label{eq:PF-ODE}
    d\vx = -t \nabla_\vx \log p_t(\vx)dt,
\end{equation}
which can be done as long as we know the time-dependent score function $\nabla_\vx \log p_t(\vx)$.
This score function is learned by a neural network $\vs_{\theta}(t, \vx)$ with parameters $\theta$, which is trained by minimizing the denoising score matching objective \cite{vincent2011connection} for every $t$:
\begin{equation}
    \E_{\vx\sim p_{\mathrm{data}}, \veps \sim \gN(0, I)}
    ||\vs_\theta(t, \vx+t\veps) + \frac{\veps}{t}||^2.
\end{equation}
By Tweedie's formula \cite{efron2011tweedie},
\begin{equation}
\label{eq:tweedie}
    D(t, \vy) = \vy + t^2\nabla_\vy \log p_t(\vy),
\end{equation}
this objective can be rewritten as learning an optimal denoiser $D_\theta(t, \vx^\mathrm{noisy})$ that aims to predict the original data $\vx$ from the corrupted data $\vx^\mathrm{noisy}= \vx+t\veps$,
\begin{equation}
    \E_{\vx\sim p_{\mathrm{data}}, \veps \sim \gN(0, I)}
    ||D_\theta(t, \vx+t\veps) - \vx||^2.
\end{equation}
A diffusion model can be easily extended to conditional distributions by simply providing a conditioning vector $\vz$ as an extra input to the network.
We now define the next-position distribution as a conditional diffusion model whose target distribution is $\nextx(\vx_{i}|\vz_{i})$, giving the following objective for learning the next-position distribution:
\begin{equation}
\label{eq:dsm}
    \E_{\vx_{i}\sim \nextx, \veps \sim \gN(0, I)}
    ||D_\theta(t, \vx_{i}+t\veps, \vz_{i}) - \vx_{i}||^2,
\end{equation}
which is a restatement of DiffLoss \cite{li2024autoregressive}.
We then predict $\hat{\vx}_i=D_\theta(t,\vx_i^\mathrm{noisy},\vz_i)$ using a Diffusion MLP (DiffMLP) with adaptive layer normalization \cite{perez2017film, peebles2022scalable}, zero-initialization \cite{peebles2022scalable}, and residual connections \cite{he2015deep}:
\begin{equation}
    \hat{\vx}_{i} = \mathrm{DiffMLP}\left(
    \mathrm{Fourier}(t)
    + \mathrm{Lin}(\vx_i^{\mathrm{noisy}})
    + \mathrm{Lin}(\mathrm{Fourier}(\vx_i^{\mathrm{noisy}}))
    + \mathrm{Lin}(\vz_i) \right).
\end{equation}
During training, once the conditioning vector $\vz_{i}$ has been constructed, we independently sample 4 timesteps $t$ to expand the batch size used for training the DiffMLP.
Once the denoiser is trained, one now has access to the score through \Cref{eq:tweedie}, and can sample $\nextx$ by drawing random noise and integrating \Cref{eq:PF-ODE} from time $t=T$ to $t=0$, using $\ndiff$ discretized timesteps.
See \Cref{app:karras} for further details on sampling, preconditioning the neural network, and timestep-weighting during training.

Molecules have translation, rotation, and permutation symmetries.
To avoid overfitting to particular orientations, we apply simple data augmentation with random rotations and random translations of up to 3Å during training.
We first center all training examples so that their center-of-mass is at the origin.
We treat molecules as ordered sequences of atoms, and we inherit the ordering of atoms as listed in the \texttt{.xyz} file.
This ordering is likely to have originated from how the original 3D structure was initialized from SMILES, or drawn by hand, which importantly provides a localized ordering of atoms.

The now-defined architecture has several characteristics to note:
It relies on standard deep learning components, such as transformers and MLPs, which have optimized hardware implementations such as FlashAttention \cite{dao2023flashattention} and optimized kernels by \texttt{torch.compile} \cite{ansel2024pytorch}.
It provides dense training supervision on every atom type and position, owing to the ability of causal transformers to efficiently batch sequences across length, while simultaneously allowing batching across multiple timesteps for training the DiffMLP.
It separates the concerns of generative modelling into modelling quadratic-scaling atom \emph{interdependence} with transformers, and modelling \emph{individual} next-type/position distributions using MLPs.
The generation cost for \name{} is $O(n)$ transformer forward passes (one per atom) and $O(n \ndiff)$ coordinate updates using a MLP.
In contrast, all-atom diffusion would require $\ndiff$ passes through a transformer or similarly expensive architecture.
This architectural distinction enables significantly faster sampling for \name{}, especially on small molecules.
\name{} operates in data-space and does not require learning a separate tokenizer \cite{liu2024bio2token}.
Thus, \name{} accepts any 3D structure as input and generates arbitrary-size output, which enables flexible use for downstream tasks such as hydrogen decoration and scaffold completion.

The causal transformer blocks are identical to GPT-2 \cite{radford2019language}, using the implementation of nanoGPT \cite{nanoGPT}, except we apply qk-layernorm along the head dimension for training stability \cite{chowdhery2023palm,wortsman2023small}, and we do not apply a final LayerNorm before output projection.
Biases are also disabled in transformer blocks.
For an architecture with $L$ transformer blocks, the first $L/2$ blocks are used for predicting $\vh_i$, while the second $L/2$ blocks are used for predicting $\vz_i$.
To reduce padding, we use sequence packing \cite{krell2021efficient}.

\vspace{-0.5em}
\subsection{Likelihood}

In diffusion models, the change-of-variables formula \cite{chen2018neural} can be used to compute the log-likelihood $\log p_0(x_0)$ for a given data point $x_0$:
\vspace{-0.5em}
\begin{equation}
\log p_0(x_0) = \log p_T(x_T) + \int_0^T \nabla \cdot \vs_\theta(t, x_t) \, dt
\end{equation}
Computing $\nabla \cdot \vs_\theta(t, x_t)$, which is the divergence (trace-Jacobian) of the score function, requires computing $d$ vector-Jacobian products, where $d$ is the dimensionality of the data.
In pure diffusion models, this is expensive because $d$ is large (e.g. $d=3n \approx 132$ for GEOM with an average of 44 atoms).
Therefore, most approaches resort to estimating log-likelihood via the ELBO or by approximating the divergence term using the Hutchinson trace estimator \cite{hutchinson1989stochastic}.
However, since the DiffMLP operates on 3-dimensional data, it is tractable to compute exact log-likelihood for each atom position by explicitly computing the full $3 \times 3$ Jacobian.

\noindent
  \begin{minipage}{\textwidth}
    \centering
    \captionof{table}{Sample quality of unconditionally generated molecules from QM9 by validity and uniqueness when generating 10,000 examples. Results for \name{} are means and standard deviations across 3 evaluation runs. \name{} uses $\ndiff=60$. Results on xyz2mol taken from \cite{gao2024tokenizing}, other results from respective works or *from our own evaluation. Higher is better, except for NLL.}
\begin{tabular}{l|cccccccc}
\toprule
\multirow{2}{*}{} & \multicolumn{1}{c}{atom} & \multicolumn{1}{c}{mol} & \multicolumn{1}{c}{lookup} & \multicolumn{1}{c}{lookup} & \multicolumn{1}{c}{xyz2mol} & \multicolumn{1}{c}{xyz2mol} & \multirow{2}{*}{NLL ($\downarrow$)}\\
 & \multicolumn{1}{c}{stable} & \multicolumn{1}{c}{stable} & \multicolumn{1}{c}{valid} & \multicolumn{1}{c}{valid$\times$uniq} & \multicolumn{1}{c}{valid} & \multicolumn{1}{c}{valid$\times$uniq} & \\
\midrule
QM9 & 99.36 & 95.30 & 97.67 & 97.63 & 99.99 & 99.90 \\
\midrule
EDM & 98.7 & 82.0 & 91.9 & 90.7 & 86.7 & 86.0 & -110.70 \\
GeoLDM & \oldtextbf{98.9} & 89.4 & 93.8 & 92.7 & 91.3 & 90.3 & - \\
SymDiff & \oldtextbf{98.9} & 89.7 & 96.4 & \oldtextbf{94.1} & 92.8* & 91.4* & \oldtextbf{-133.79} \\
\midrule
G-SchNet & 95.7 & 68.1 & 85.5 & 80.3 & 75.0 & 72.5 & - \\
Symphony & 90.8 & 43.9 & 68.1 & 66.5 & 83.5 & 81.8 & - \\
Mol-StrucTok & 98.5 & 88.3 & \oldtextbf{98.0} & 83.4 & 96.7 & 82.5 & - \\
\name{} & 98.7 & \oldtextbf{90.4} & 95.7 & 90.2 & \oldtextbf{98.6} & \oldtextbf{94.0} & -97.03 \\[-4pt]
& $\spm{0.0}$ & $\spm{0.4}$ & $\spm{0.2}$ & $\spm{0.2}$ & $\spm{0.1}$ & $\spm{0.3}$ & \\

\bottomrule
\end{tabular}
    \label{qm9-table}
  \end{minipage}
  
  \vspace{1em}  %

  \begin{minipage}{\textwidth}
    \centering
    \includegraphics[width=\textwidth]{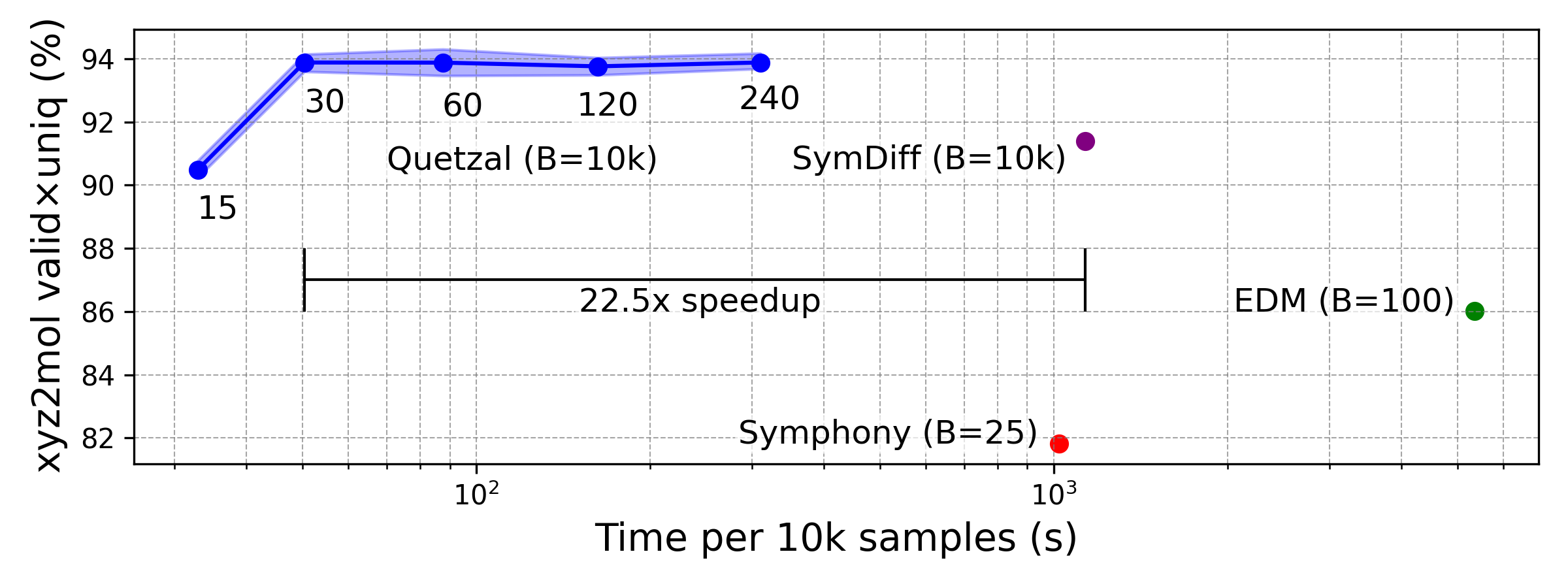}
    \captionof{figure}{xyz2mol validity$\times$uniqueness of 10k samples as a function of generation speed for QM9. $B$ is the batch size used for generation. We show the largest batch size that fits on a single A100 40 GB GPU. For \name{}, the text annotation is the number of diffusion steps used per atom. Error bars show min/max over 5 evaluations. See Appendix \Cref{speed-batch} for generation speed vs batch size.}
    \label{speed-valid}
  \end{minipage}

\section{Experiments}

\subsection{Molecular generation}

We train \name{} on unconditional 3D molecular generation from the QM9 \cite{ramakrishnan2014quantum} and GEOM-DRUGS \cite{axelrod2022geom} datasets (abbreviated as GEOM).
We follow the train/val/test splits of \citet{hoogeboom2022equivariant}, as well as their evaluation protocol of generating 10,000 molecules and assessing atom stability, molecule stability, and validity and uniqueness via bond lookup tables.
We also assess validity and uniqueness via xyz2mol as introduced by \citet{daigavane2023symphony}.
Finally, we assess negative log-likelihood (NLL) of the test set according to each model.
We implement \name{} using a 12-layer transformer (12 attention heads, hidden size 768, 86M parameters) and a 6-layer DiffMLP (hidden size 1536, 79M parameters), resulting in 165M total parameters.
More details on metrics, evaluation, generated samples, and ablation studies are in \Cref{app:exp}.

\textbf{Baselines}. We compare to equivariant diffusion models EDM \cite{hoogeboom2022equivariant}, GeoLDM \cite{xu2023geometric}, and SymDiff \cite{zhang2024symdiff}.
EDM and GeoLDM use equivariant graph neural networks (EGNNs) \cite{satorras2021n}, whereas SymDiff uses a permutation-equivariant diffusion transformer that scalably incorporates rotation equivariance via stochastic symmetrization.
We also compare to autoregressive models G-SchNet \cite{gebauer2019symmetry} and Symphony \cite{daigavane2023symphony}, which rely on equivariant, relative predictions of the next atom position,
as well as Mol-StrucTok \cite{gao2024tokenizing}, which tokenizes 3D structures into an SE(3)-invariant line notation for language modeling.
All autoregressive baselines discretize 3D space.

\subsection{QM9 generation results}

\textbf{Validity and uniqueness.} \name{} achieves strong sample quality on QM9, outperforming prior autoregressive methods in both xyz2mol and lookup table metrics (\Cref{qm9-table}), and surpassing pure diffusion models in xyz2mol metrics.
Interestingly, \name{} exhibits signs of overfitting as evidenced by high validity but reduced validity$\times$uniqueness.
\name{} also obtains a poor estimate of test-set log-likelihood, despite generating high-quality samples.
These observations may stem from \name{} overfitting to the fixed atom orderings seen during training.

\textbf{Generation efficiency.} 
\name{} generates molecules significantly faster than all baselines, despite having the most parameters.
\Cref{speed-valid} shows the tradeoff between sample quality and generation time on a single A100 40GB GPU, where each model is run with the largest batch size that fits in memory. At $\ndiff = 30$, \name{} achieves a 22.5$\times$ speedup over SymDiff while obtaining better xyz2mol validity$\times$uniqueness.
Although recent samplers \cite{song2020denoising, lu2022dpm, karras2022elucidating} can reduce inference steps for diffusion models, matching \name{}'s speed would require reducing SymDiff's 1000 steps to fewer than 44 — while preserving quality.
Importantly, \name{} could also benefit from such sampler improvements.
In Appendix \Cref{speed-batch}, we show that generation throughput also scales well with batch size.

The speed can be attributed to several factors:
(1) Whereas pure diffusion models require a dense pass through the transformer on every timestep, \name{} only calls the transformer once per new atom, and instead relies on cheaper passes through the MLP for diffusion.
(2) The number of diffusion steps is also largely reduced by using the Heun sampler and geometrically-spaced timesteps proposed by \citet{karras2022elucidating}.
(3) Forward passes are cheaper because the optimized performance of FlashAttention \cite{dao2023flashattention} is much faster than expensive message-passing steps of EGNN \cite{satorras2021n} or tensor products for Symphony.

\subsection{GEOM generation}

\name{} is, to our knowledge, the first autoregressive model demonstrated on the large and diverse GEOM dataset.
We compare to diffusion-based baselines including GCDM \cite{morehead2024geometry}.
\name{} again achieves generation quality approaching that of diffusion models (\Cref{geom-table}), but with much faster sampling:
\name{} ($\ndiff=120$) requires 11.9 minutes for 10k samples, whereas EDM requires 1,533 minutes (128$\times$ speedup) and GCDM requires 683 minutes (57$\times$ speedup).
Interestingly, \name{} achieves state-of-the-art NLL on GEOM, despite underperforming on QM9.
We hypothesize that this is due to random splitting of GEOM:
The dataset includes up to 30 conformers per molecule, so most molecules in the test set have conformers that are seen in the training set.
This also explains why lookup validity$\times$uniqueness of the original training set is low.
We show uncurated samples in Appendix \Cref{geom_uncurated}.

\begin{table}[t]
\centering
\caption{Sample quality of unconditionally generated molecules from GEOM by validity and uniqueness. \name{} uses $\ndiff=120$ for generation and $\ndiff=60$ for NLL. *We assume uniqueness is 100\%.}
\label{geom-table}
\begin{tabular}{l|ccccc}
\toprule
\multirow{2}{*}{} & \multicolumn{1}{c}{atom} & \multicolumn{1}{c}{lookup} & \multicolumn{1}{c}{lookup} \\
 & \multicolumn{1}{c}{stable} & \multicolumn{1}{c}{valid} & \multicolumn{1}{c}{valid$\times$uniq} & NLL ($\downarrow$) \\
\midrule
GEOM & 86.5 & 99.9 & 69.5 &\\
\midrule
EDM & 81.3 & 92.6 & 92.6* & -137.1\\
GeoLDM & 84.4 & 99.3 & 99.3* & - \\
GCDM & 89.0 & 95.5 & 95.5* & -234.3 \\
SymDiff & 86.2 & 99.3 & 99.3* & -301.21 \\
\midrule
\name{} & 86.7$\spm{0.0}$ & 95.6$\spm{0.1}$ & 95.3$\spm{0.2}$ & -313.63\\

\bottomrule
\end{tabular}
\end{table}

\subsection{Hydrogen decoration}

Because \name{} generates atoms sequentially, with hydrogens typically last, it can be applied with no additional training to decorate 3D structures with missing hydrogens.
This task is useful for adding hydrogens to 3D structures from X-ray crystallography, which often lack resolved hydrogens due to low electron density \cite{muller2009practical}.
Typically, this task is performed with specialized cheminformatics software.
It is unclear how to apply pure diffusion models to the task of hydrogen decoration, as they require specifying a fixed size.
Therefore, we compare to tools such as the crystallography toolbox Olex2 \cite{dolomanov2009olex2}, and OpenBabel + Hydride \cite{o2011open, kunzmann2022adding}, which first infers bonds then adds hydrogens in 3D.
We also compare to Symphony, an autoregressive model trained on QM9.
We test this task by stripping hydrogens from the test set of QM9, and evaluate the accuracy in adding hydrogens back.

As metrics, we check whether each method adds the correct number of hydrogens.
If the number of hydrogens is correct, we calculate the root-mean-squared deviation (RMSD) for just the hydrogen atoms, and check whether it satisfies thresholds of 0.5, 0.1, and 0.05 Å.
Because hydrogens can be added in any order, we first assign permutations between predicted and ground truth by solving a linear assignment problem (Hungarian algorithm) on Euclidean distances.
Results are in \Cref{hydrogen-addition-table}.

\name{} predicts hydrogens with high accuracy.
However, \name{}'s hydrogen predictions are sensitive to atom ordering.
\name{} is able to solve this task because in QM9, hydrogens appear last in the \texttt{.xyz} files.
If the bare molecule without hydrogens is reordered, \name{}'s performance degrades significantly, as the prefix becomes out-of-distribution.
Additional results are in \Cref{app:hdeco}.

\begin{table}
\caption{Method performance on adding hydrogens onto bare molecules from the test set of QM9. All results are from our own evaluation. *Checkpoint appears to be undertrained, see \Cref{app:hdeco}.}
\label{hydrogen-addition-table}
\centering
\begin{tabular}{l|cccc}
\toprule

\multirow{2}{*}{Method} & \multicolumn{1}{c}{Correct} & \multicolumn{3}{c}{\% $<$ RMSD Å} \\
& \multicolumn{1}{c}{Num H} & \multicolumn{1}{c}{0.5} & \multicolumn{1}{c}{0.1} & \multicolumn{1}{c}{0.05} \\
\midrule
Olex2 & 62.7 & 57.1 & \hfill 7.8 & \hfill 0.1 \\
OpenBabel+Hydride & 88.4 & 79.0 & 42.9 & 12.7 \\
Symphony* & 46.9 & 43.6 & 34.9 & 23.8 \\
\name{} & \oldtextbf{99.8} & \oldtextbf{99.5} & \oldtextbf{94.1} & \oldtextbf{90.4} \\

\bottomrule
\end{tabular}
\end{table}

\begin{figure}
    \centering
    \centerline{\includegraphics[width=\textwidth]{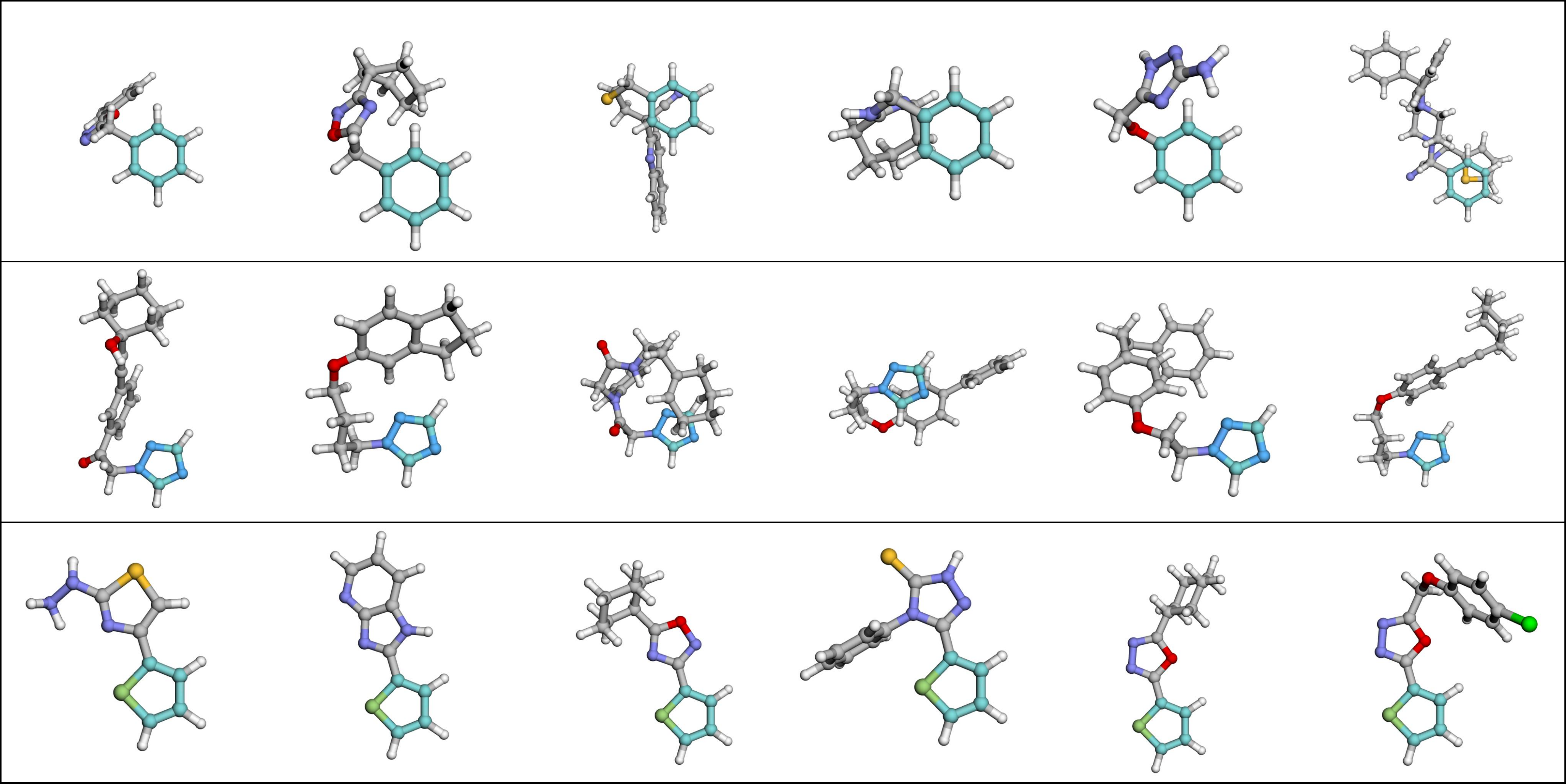}}
    \caption{Selected examples of scaffold completion for benzene, 1,2,4-triazole, and thiophene. Generation uses $\ndiff=120$.}
    \label{scaffold}
\end{figure}

\subsection{Scaffold completion}

Scaffold completion is naturally suited to autoregressive generation, since the prefix structure is held fixed.
We demonstrate completions for benzene, 1,2,4-triazole, and thiophene scaffolds in \Cref{scaffold}.
These are qualitative results; we defer quantitative evaluation and comparison \cite{xie2024diffdec} to future work.
We note that, like hydrogen decoration, scaffold completion is sensitive to the initial scaffold configuration in terms of its atom ordering, center of mass, and orientation.
However, this can be useful to steer how the scaffold is completed.

\section{Discussion}

Our architecture is simple: it does not require a separate tokenizer or autoencoder, does not model bonds explicitly, does not predict a focal atom or relative coordinates, and does not architecturally consider permutation, translation, or rotation symmetries.
Instead, we rely on a standard transformer backbone equipped with Diffusion Loss, a simple method for autoregressive generation of per-token continuous coordinates \cite{li2024autoregressive}.
In doing so, we create a model that is simple to implement, trainable at scale, and fast to sample from.

Despite these advantages, the model has limitations.
It is not permutation-invariant.
Removing positional encodings does not confer permutation symmetry, as the generation order still induces a learned ordering \cite{haviv2022transformer, kazemnejad2024impact}.
This dependency on generation order also constrains generalization—e.g., the model performs poorly under random atom orderings and cannot generalize to molecules with more atoms than seen during training.
In ablations, we show that randomly permuting the atom ordering leads to poor performance (\Cref{app:ablation}), similar to how training autoregressive Sudoku solvers is highly dependent on generation order \cite{shah2024causal}.
Therefore, we need to know the best order in which to generate atoms.
Additionally, like other autoregressive models, our method is susceptible to error accumulation \cite{lecun2023large} or failures in teacher-forcing \cite{bachmann2024pitfalls}.

Future work may explore training and decoding strategies that reduce dependence on or infer atom generation order, such as masked diffusion \cite{kim2025train}.
In addition, future work can also exploit the fact that autoregressive models accept \emph{arbitrary-size input} and generate \emph{arbitrary-size output}, which can provide extremely flexible conditioning, especially in the context of text \cite{gruver2024fine}.
Autoregression could also allow the model to reason for many continuous-valued tokens using chain-of-thought \cite{wei2022chain,hao2024training}.
Finally, tractable exact likelihood computation enables importance sampling for Boltzmann generators \cite{noe2019boltzmann, klein2023equivariant, klein2024transferable, tan2025scalable}, and could unlock new strategies for finetuning models on reward functions.

This work advances molecular generative modeling, with potential to accelerate molecular simulation and discovery.
These capabilities can benefit drug discovery, materials design, and the development of sustainable technologies—areas with direct impact on public health, energy, and environmental sustainability.
At the same time, we acknowledge that these capabilities could advance the discovery of (bio)chemical weapons \cite{urbina2022dual}.
Nevertheless, the effective use or misuse of such technologies would require significant wet-lab resources and expertise.

\section{Acknowledgements}
A.H.C. thanks Marta Skreta and Lazar Atanackovic for helpful discussions.
This research was enabled in part
by computational resources provided by the Digital Research Alliance of Canada (\url{https://alliancecan.ca}) and the Acceleration Consortium (\url{https://acceleration.utoronto.ca}).
A.H.C. acknowledges the generous support of the Canada 150 Research Chairs program through A.A.-G.
A.A.-G. thanks Anders G.~Frøseth for his generous support, and acknowledges the generous support of Natural Resources Canada and the Canada 150 Research Chairs program.
This research is part of the University of Toronto’s Acceleration Consortium, which receives funding from the Canada First Research Excellence Fund (CFREF) via CFREF-2022-00042.

We acknowledge the Python community \citep{van1995python,oliphant2007python} for developing the core set of tools that enabled this work, including PyTorch \citep{paszke2019pytorch}, PyTorch Lightning \citep{Falcon_PyTorch_Lightning_2019}, RDKit \citep{greg_landrum_2023_8254217}, py3Dmol \citep{rego20153dmol}, Jupyter \citep{kluyver2016jupyter},
Matplotlib \citep{hunter2007matplotlib}, seaborn \citep{waskom2021seaborn}, NumPy \citep{2020NumPy-Array}, SciPy \citep{2020SciPy-NMeth}, and pandas \citep{The_pandas_development_team_pandas-dev_pandas_Pandas}.

\bibliography{refs}
\bibliographystyle{unsrtnat}

\newpage
\appendix

\begin{figure}[t]
\centering
\centerline{\includegraphics[width=\textwidth]{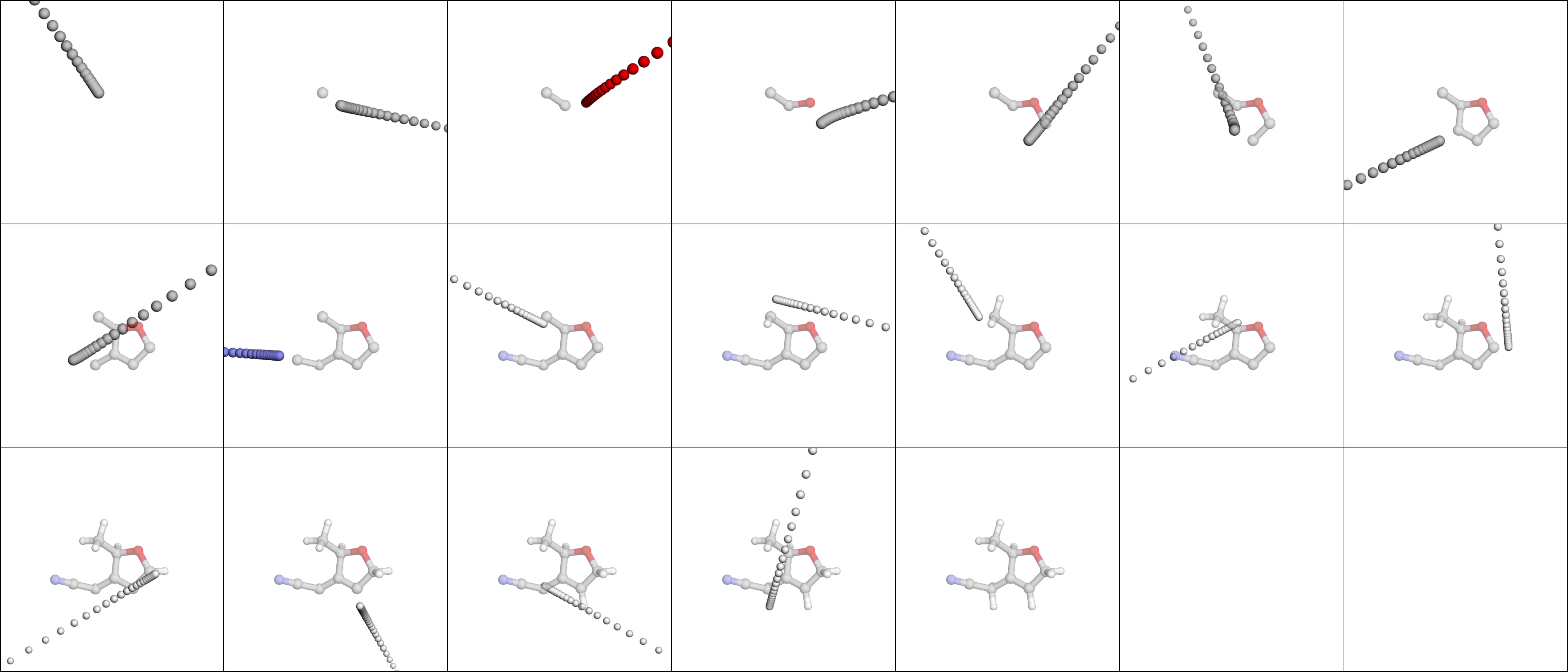}}
\caption{\name{} generates 3D molecules by iteratively predicting the next atom's discrete type and continuous 3D position. The continuous trajectories of the DiffMLP are shown at every step.}
\label{sequence}
\end{figure}

\begin{figure}
    \centering
    \centerline{\includegraphics[width=\textwidth]{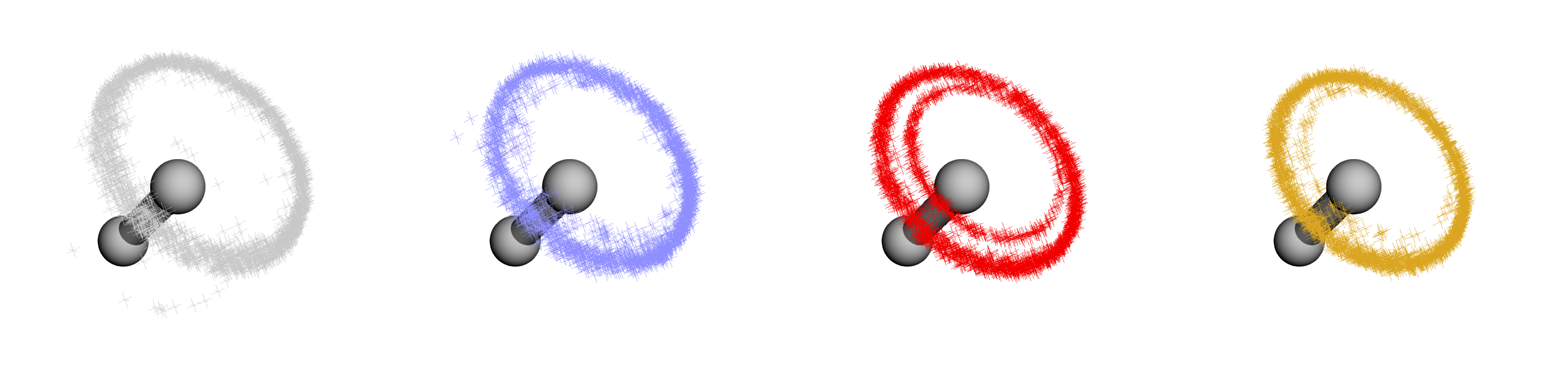}}
    \caption{Next position distributions for carbon, nitrogen, oxygen, and fluorine. The model learns to make symmetric predictions from data augmentation.}
    \label{2atom}
\end{figure}

\section{\citet{karras2022elucidating} diffusion framework}
\label{app:karras}

Following the framework of \citet{karras2022elucidating}, we precondition the neural network using the following reparameterization,
\begin{equation}
    D_\theta(t,\vx^{\mathrm{noisy}}) = \frac{\sdata^2}{t^2+\sdata^2}\vx^{\mathrm{noisy}}+\frac{t\sdata}{\sqrt{t^2+\sdata^2}}F_\theta\left(\frac{\vx^{\mathrm{noisy}}}{\sqrt{t^2+\sdata^2}},\frac{1}{4}\ln t\right),
\end{equation}
where $\sdata$ is a hyperparameter set to the standard deviation of the coordinates depending on the dataset, and $F_\theta$ is the actual DiffMLP.

During training, timesteps are sampled from a log-normal distribution $\ln t\sim \gN(-1.2, 1.2^2)$, and the denoising score matching loss of \Cref{eq:dsm} on each timestep is weighted by $(t^2+\sdata^2) / (t\sdata)^2$.
For sampling, we use the Heun integrator, which evaluates $D_\theta$ twice per integration timestep, and the discretized timesteps are geometrically spaced, given by
\begin{equation}
    t_i = \left(\smax^{1/\rho}+\frac{i}{\ndiff-1}(\smin^{1/\rho}-\smax^{1/\rho})\right)^\rho,
\end{equation}
where $\ndiff$ is the number of diffusion steps, $\rho=7$, $\smin=10^{-4}$, and $\smax=80$.

\section{Experimental details}
\label{app:exp}

We use the same train/val/test splits as \citet{hoogeboom2022equivariant}, which contain 10,000/17,748/13,083 examples for QM9 and 5,538,014/692,251/692,251 examples for GEOM.

We train models on QM9 for 2000 epochs.
We use sequence packing, using the Longest-pack-first histogram-packing algorithm \cite{krell2021efficient}.
Before training, we pack all examples into sequences of size 128, enforcing a maximum of 6 examples per pack.
We then batch packs together by concatenating across the length dimension, with a batch size of 180 packs.
This procedure of batching packs was necessary for keeping uniform the number of examples per pack, which was vital for efficient and stable training convergence.
We train for 2000 epochs (188k steps) on a single A100 40GB GPU, which took a wall-time of 21 hours.
We do document masking using FlexAttention \cite{dong2024flex}, preventing the model from attending to other examples in the batched pack.

We do gradient clipping to a norm of 1.0.
We train with AdamW \cite{loshchilov2017decoupled} using a learning rate of $4\times10^{-4}$, $\beta_1=0.9$, $\beta_2=0.95$, and weight decay = $10^{-5}$.
We maintain an exponential moving average of the parameters with decay rate 0.999.
We use $\sigma_\mathrm{data}=1.4$ for QM9, and $\sigma_\mathrm{data}=2.5$ for GEOM.

Fourier encodings were important for efficient learning and are used in three different parts of the architecture:
\begin{enumerate}
    \item For embedding coordinates in the transformer, each element of a vector of size 3 is mapped to 256 Fourier channels with bandwidth $b=20$, before flattening to size 768.
    \item For embedding coordinates in the DiffMLP, each element of each vector of size 3 is mapped to 512 Fourier channels with bandwidth $b=20$, before flattening to size 1536.
    \item For embedding timestep in the DiffMLP, a scalar is mapped to $w$ Fourier channels with bandwidth $b=1$, where $w$ is the width of the DiffMLP.
\end{enumerate}
We use the magnitude-preserving Fourier encodings proposed by \citet{karras2024analyzing}.
A scalar $x$ is mapped to a vector of Fourier features via $x\mapsto\sqrt{2}\cos \big( 2\pi (bf_i x + \varphi_i) \big)$, where $b$ is the bandwidth, and frequencies $f_i \sim \gN(0,1)$ and phases $\varphi_i\sim\gU[0,1]$ are randomly initialized constants.

For GEOM, we train with 4 A100 40GB GPUs for 201 epochs (734k steps).
We use a learning rate of $2\times 10^{-4}$ per GPU. 
We pack all examples into sequences of size 512, enforcing a maximum of 10 examples per pack, and use a batch size of 40 packs per batch.

\begin{figure}[ht]
\centering
\centerline{\includegraphics[width=\textwidth]{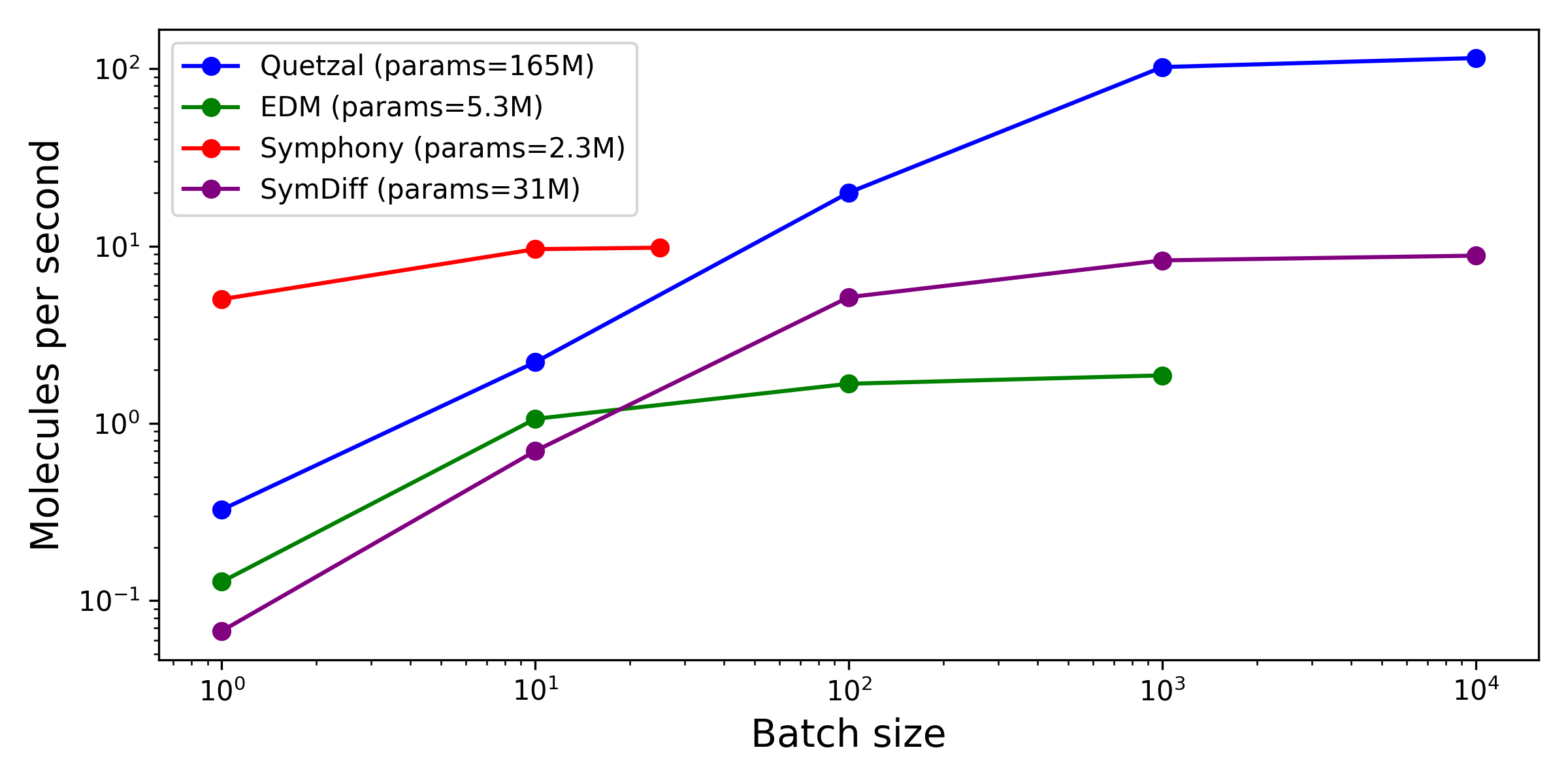}}
\caption{Generation speed of QM9 examples as a function of batch size on a single A100 40GB GPU. Despite having over 5$\times$ as many parameters as baselines, \name{} scales to large batch sizes at inference time, enabling fast amortized generation.}
\label{speed-batch}
\end{figure}

\subsection{Metrics}
\label{app:metrics}

\citet{garcia2021n} introduce several metrics for evaluating the quality of generated 3D molecules.
They define a lookup table of allowed bond lengths, with thresholds tuned to maximize the validity of each dataset.
A molecule is assigned bonds using this lookup table, and then the valency of each atom is checked.
An atom is stable if it has the correct valency.
A molecule is stable if all of its atoms are stable.
Atom stability is the proportion of generated atoms which are stable.
Molecule stability is the proportion of generated molecules which are stable.
A molecule is valid if its assigned bonds can be parsed by RDKit without failure.
Validity is the proportion of generated examples which are valid.
If the molecule can be parsed by RDKit, then it can be turned into a SMILES string.
Uniqueness is calculated as the number of unique, generated SMILES strings divided by the number of generated molecules.

We use RDKit's xyz2mol \cite{kim2015universal}, specifically we use \texttt{rdkit==2023.03.3} with \texttt{rdDetermineBonds.DetermineBonds(mol, charge=0)}.
A molecule is valid if this function passes without error and the resulting molecule can be turned into a SMILES string.
We found that this version of RDKit reproduces the results of \cite{daigavane2023symphony}, and determines 99.99\% of the QM9 training set to be valid, whereas later versions of RDKit only determines 94.78\% to be valid.

We estimate the first row of \Cref{geom-table} by computing metrics for 200k random examples from the training set of GEOM.

\clearpage

\subsection{Architecture ablation}
\label{app:ablation}

\begin{figure}[!htbp]
\centering
    \centerline{\includegraphics[width=\textwidth]{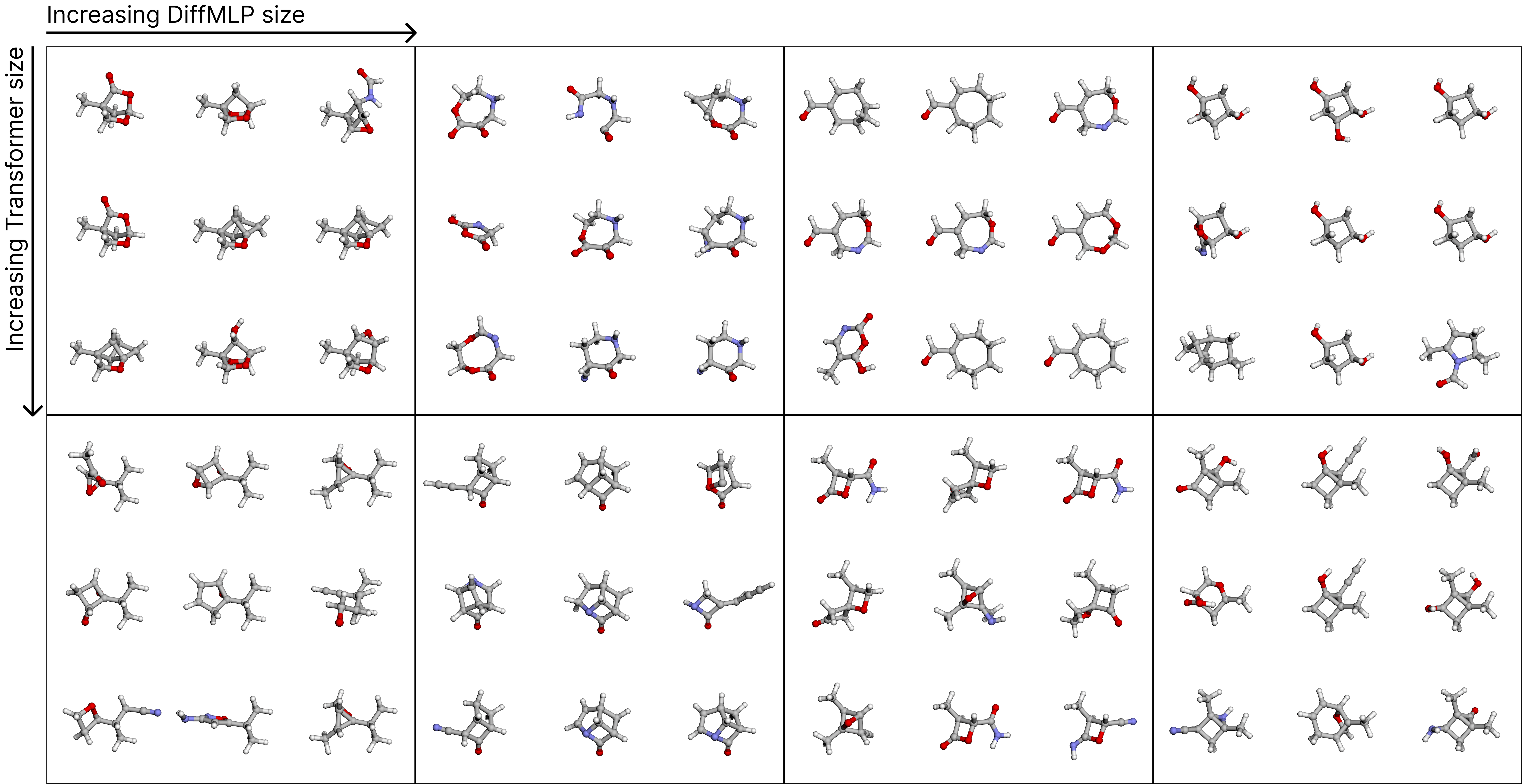}}
    \caption{Generated samples using the same random generation seed for different sized models. $\ndiff=30$ diffusion steps are used for each atom. Different models converge to similar molecules. Both the transformer and DiffMLP are important in controlling structure. Model sizes in \Cref{ablate-size}.}
\label{seeded-ablation}
\end{figure}

\begin{table}[!htbp]
\caption{Ablation of transformer and DiffMLP size. $W$ is the transformer width, $H$ is the number of heads, and $L$ is the number of layers. $w$ is the DiffMLP width. Results show mean and standard deviation across 3 evaluation runs.}
\label{ablate-size}
\centering
\begin{tabular}{l|l|ccccccc}
\toprule
Transformer & MLP & atom & mol & lookup & lookup & xyz2mol & xyz2mol \\
size & width & stable & stable & valid & valid$\times$uniq & valid & valid$\times$uniq \\
\midrule

$W=512$ & $w=512$ & 96.0$\spm{0.1}$ & 73.8$\spm{0.4}$ & 87.5$\spm{0.2}$ & 84.4$\spm{0.2}$ & 95.8$\spm{0.1}$ & 91.6$\spm{0.2}$ \\
$H=8$ & $w=1024$ & 97.4$\spm{0.1}$ & 81.6$\spm{0.5}$ & 91.6$\spm{0.2}$ & 88.1$\spm{0.3}$ & 98.2$\spm{0.1}$ & 94.0$\spm{0.3}$ \\
$L=8$ & $w=1536$ & 97.7$\spm{0.0}$ & 83.4$\spm{0.2}$ & 92.6$\spm{0.2}$ & 88.9$\spm{0.1}$ & 98.4$\spm{0.1}$ & 94.0$\spm{0.0}$ \\
\midrule
$W=640$ & $w=512$ & 97.1$\spm{0.1}$ & 80.6$\spm{0.4}$ & 91.1$\spm{0.4}$ & 87.3$\spm{0.5}$ & 96.8$\spm{0.2}$ & 92.2$\spm{0.2}$ \\
$H=10$ & $w=1024$ & 97.6$\spm{0.0}$ & 82.9$\spm{0.3}$ & 92.5$\spm{0.1}$ & 88.7$\spm{0.1}$ & 98.4$\spm{0.1}$ & 93.9$\spm{0.2}$ \\
$L=10$ & $w=1536$ & 98.0$\spm{0.1}$ & 85.7$\spm{0.6}$ & 93.8$\spm{0.4}$ & 89.8$\spm{0.4}$ & 98.9$\spm{0.1}$ & 94.0$\spm{0.2}$ \\
\midrule
$W=768$ & $w=512$ & 96.7$\spm{0.1}$ & 78.6$\spm{0.4}$ & 90.3$\spm{0.1}$ & 85.9$\spm{0.1}$ & 97.1$\spm{0.0}$ & 91.7$\spm{0.2}$ \\
$H=12$ & $w=1024$ & 97.9$\spm{0.0}$ & 85.8$\spm{0.2}$ & 93.6$\spm{0.2}$ & 89.3$\spm{0.2}$ & 98.4$\spm{0.1}$ & 93.5$\spm{0.3}$ \\
$L=12$ & $w=1536$ & 98.3$\spm{0.0}$ & 87.6$\spm{0.3}$ & 94.7$\spm{0.2}$ & 90.1$\spm{0.1}$ & 99.1$\spm{0.0}$ & 94.0$\spm{0.1}$ \\
\midrule
\multicolumn{2}{c|}{with atom permutations} & 82.3$\spm{0.1}$ & 25.9$\spm{0.2}$ & 63.1$\spm{0.5}$ & 62.5$\spm{0.4}$ & 80.7$\spm{0.2}$ & 80.1$\spm{0.3}$ \\
\multicolumn{2}{c|}{w/o translations \& rotations} & 84.8$\spm{0.1}$ & 22.0$\spm{0.1}$ & 48.5$\spm{0.5}$ & 47.3$\spm{0.5}$ & 63.0$\spm{0.3}$ & 60.9$\spm{0.4}$ \\

\bottomrule
\end{tabular}
\end{table}

\FloatBarrier

\begin{figure}[H]
    \centering
    \centerline{\includegraphics[width=\textwidth]{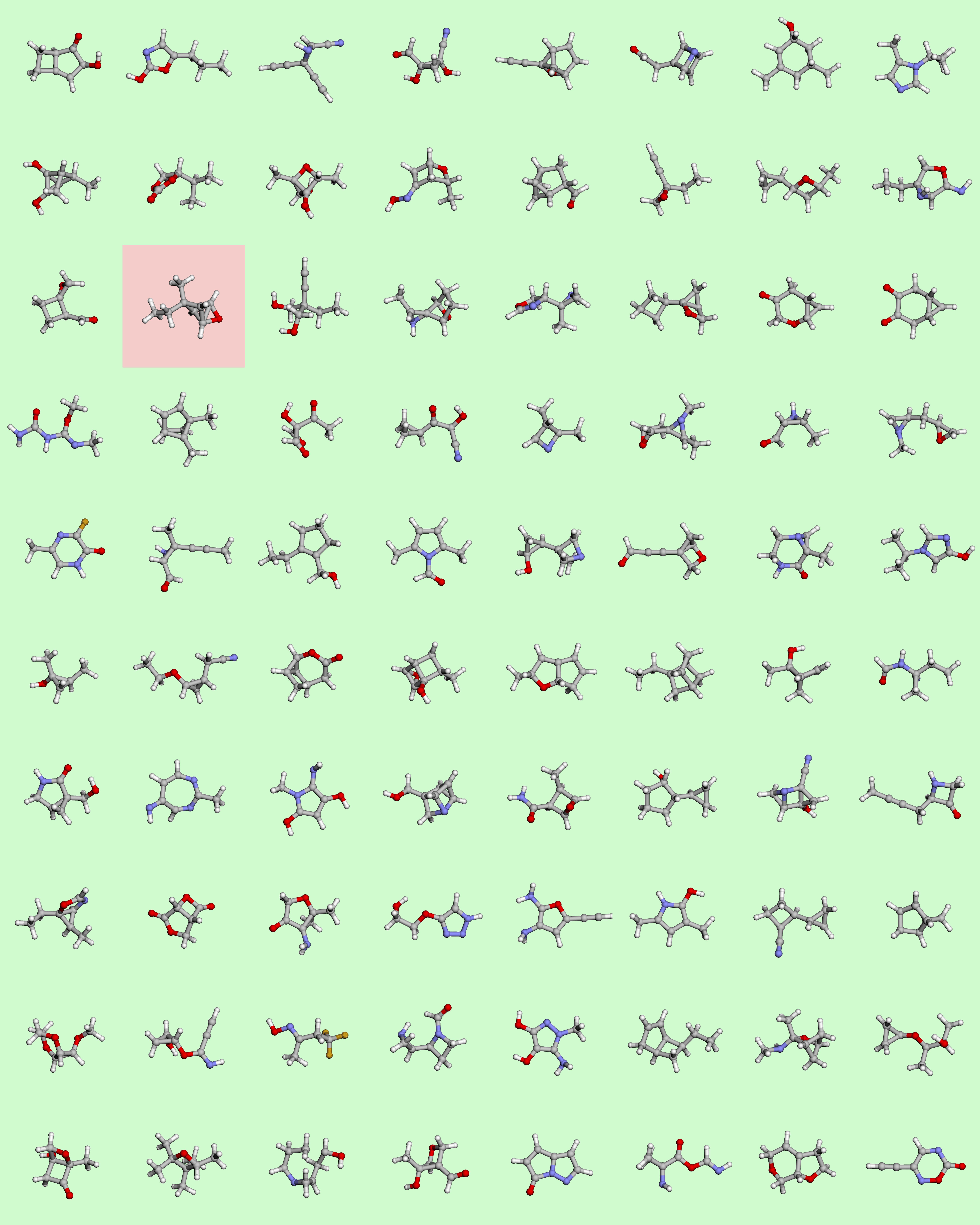}}
    \caption{Uncurated generated molecules from QM9. Green/red indicates valid/invalid by xyz2mol.}
    \label{qm9_uncurated}
\end{figure}

\begin{figure}[H]
    \centering
    \centerline{\includegraphics[width=\textwidth]{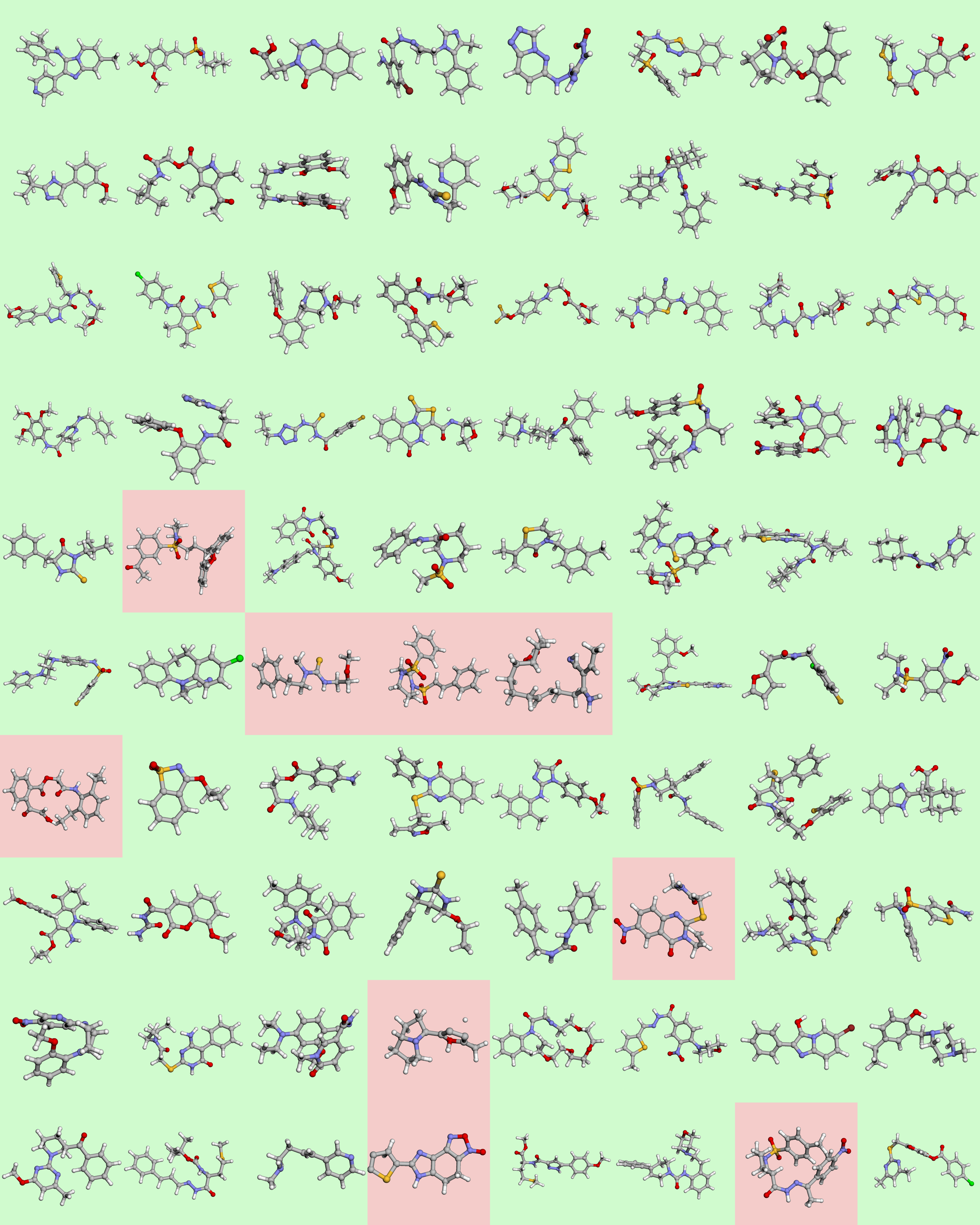}}
    \caption{Uncurated generated molecules from GEOM. Green/red indicates valid/invalid by xyz2mol.}
    \label{geom_uncurated}
\end{figure}

\FloatBarrier

\subsection{Hydrogen decoration}
\label{app:hdeco}

\begin{figure}[H]
\centering
\centerline{\includegraphics[width=\textwidth]{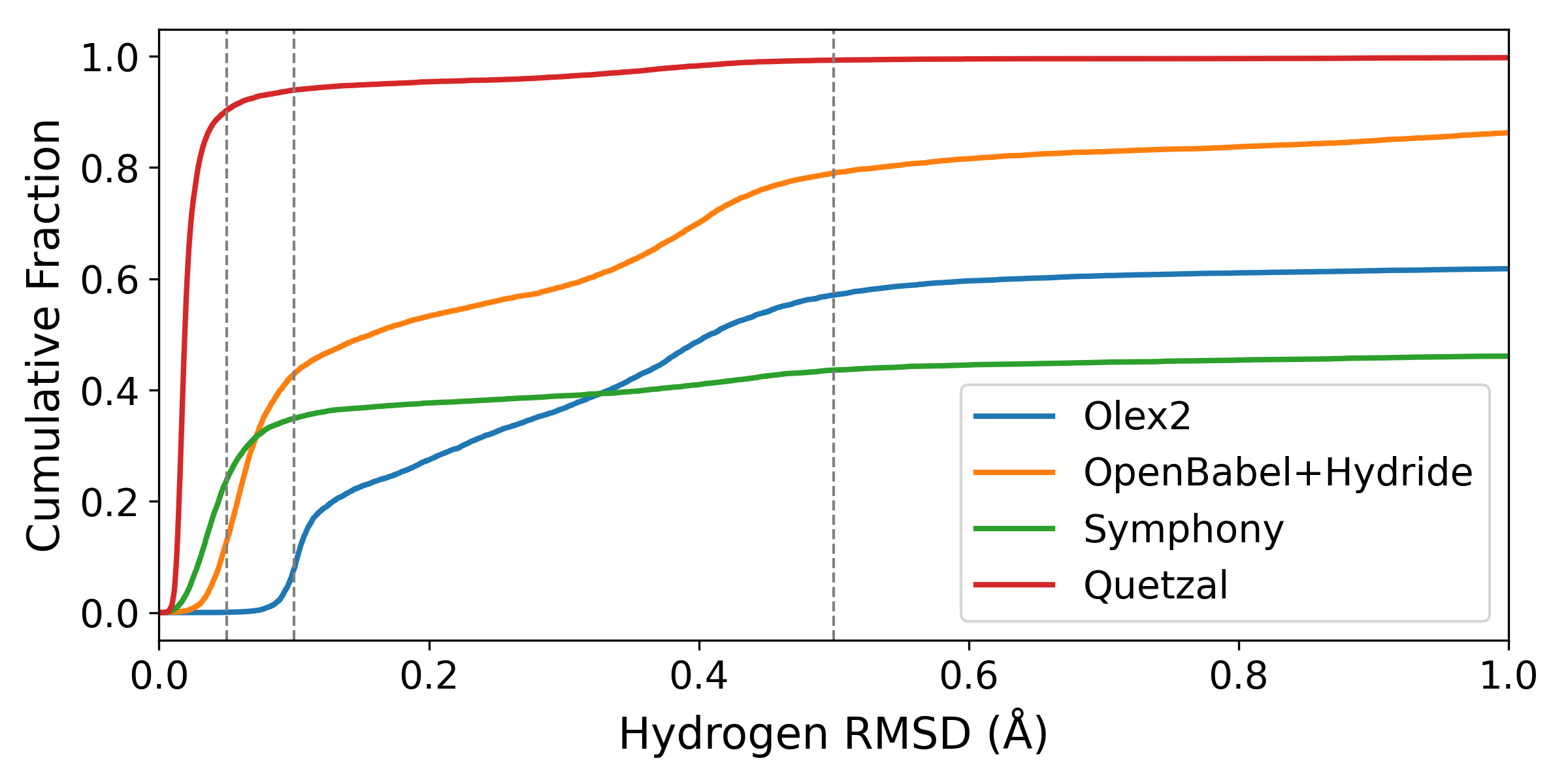}}
\caption{Cumulative distribution functions of RMSD after decorating bare molecules from the test set of QM9. \name{} adds hydrogens with very low RMSD for a large majority of the test set. Adding an incorrect number of hydrogens is treated as $\mathrm{RMSD}=\infty$. The vertical dotted lines are the thresholds 0.5, 0.1, 0.05 Å as shown in \Cref{hydrogen-addition-table}. The checkpoint for Symphony appears to be undertrained: \url{https://github.com/atomicarchitects/symphony/blob/3f2c6a7f7983877f4a5f2a0a71328b29bdc553cf/tutorial/workdir/checkpoints/params_best.pkl}}
\label{hdeco-rmsd}
\end{figure}

\begin{figure}[H]
\centering
\centerline{\includegraphics[width=\textwidth]{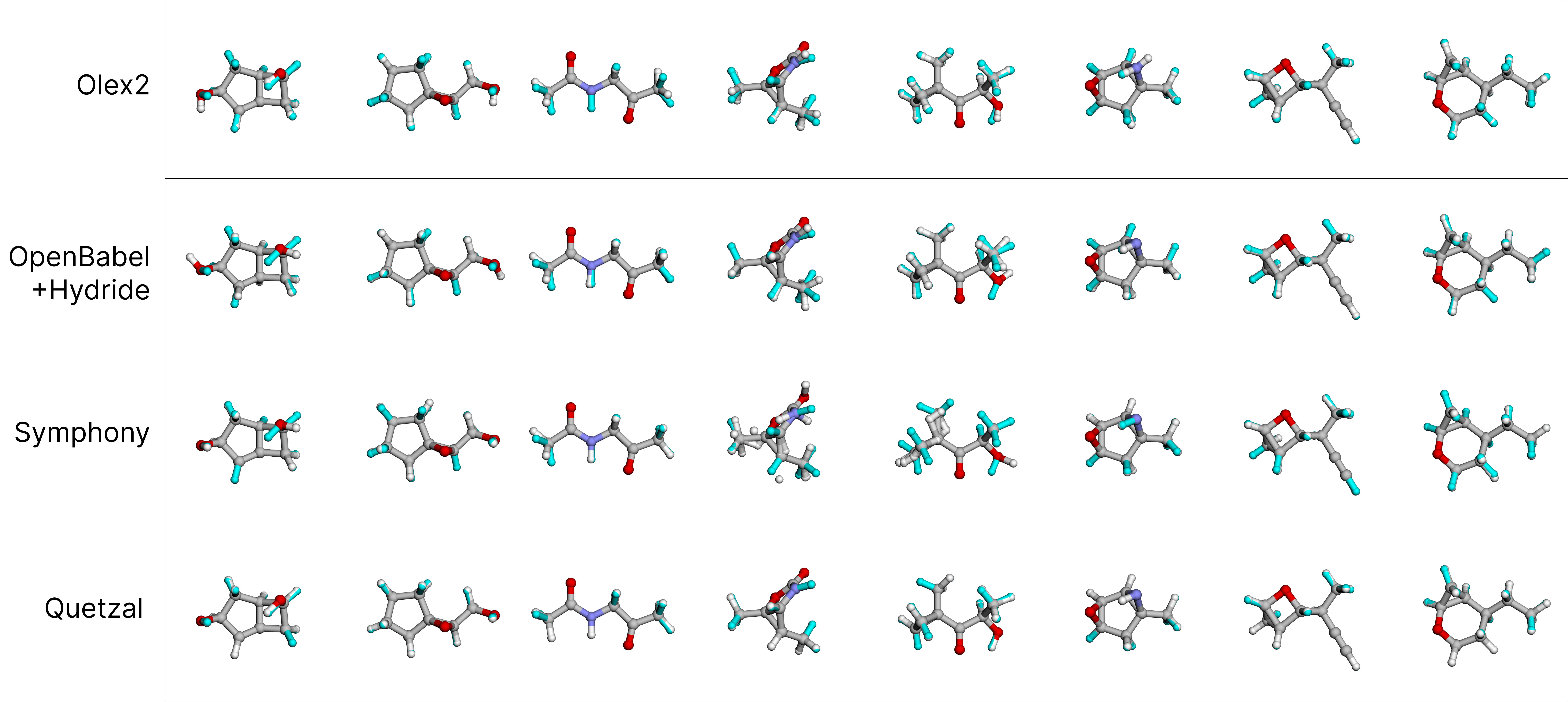}}
\caption{Comparison of methods for adding hydrogens in 3D. The ground truth is displayed in cyan. Accurate hydrogen placement for hydroxyl and methyl groups is difficult.}
\label{hdeco-examples}
\end{figure}

\newpage
\section{Licenses}

Datasets:
\begin{itemize}
    \item QM9 \cite{ramakrishnan2014quantum}: The license status is unclear
    \item GEOM \cite{axelrod2022geom}: CC0 1.0 Universal
\end{itemize}

Models:
\begin{itemize}
    \item EDM \cite{hoogeboom2022equivariant}: MIT License
    \item Symphony \cite{daigavane2023symphony}: MIT License
    \item SymDiff \cite{zhang2024symdiff}: MIT License
\end{itemize}

\end{document}